# Nowcasting-Nets: Deep Neural Network Structures for Precipitation Nowcasting Using IMERG

Mohammad Reza Ehsani[1], Ariyan Zarei[2], Hoshin V. Gupta[1], Kobus Barnard[2], Ali Behrangi[1, *]

[1] Department of Hydrology and Atmospheric Sciences, The University of Arizona; Tucson, AZ, 85721

[2] Department of Computer Science, The University of Arizona; Tucson, AZ, 85721

rehsani@email.arizona.edu; ariyanzarei@email.arizona.edu; kobus@email.arizona.edu; hoshin@email.arizona.edu;

* Correspondence: behrangi@email.arizona.edu

# Abstract

Accurate and timely estimation of precipitation is critical for issuing hazard warnings (e.g., for flash floods or landslides). Current remotely sensed precipitation products have a few hours of latency, associated with the acquisition and processing of satellite data. By applying a robust nowcasting system to these products, it is (in principle) possible to reduce this latency and improve their applicability, value, and impact. However, the development of such a system is complicated by the chaotic nature of the atmosphere, and the consequent rapid changes that can occur in the structures of precipitation systems

In this work, we develop two approaches (hereafter referred to as *Nowcasting-Nets*) that use *Recurrent* and *Convolutional* deep neural network structures to address the challenge of precipitation nowcasting. A total of five models are trained using *Global Precipitation Measurement* (GPM) *Integrated Multi-satellitE Retrievals for GPM* (IMERG) precipitation data over the Eastern Contiguous United States (CONUS) and then tested against independent data for the Eastern and Western CONUS. The models were designed to provide forecasts with a lead time of up to 1.5 hours and, by using a feedback loop approach, the ability of the models to extend the forecast time to 4.5 hours was also investigated.

Model performance was compared against the Random Forest (RF) and Linear Regression (LR) machine learning methods, and also against a persistence benchmark (BM) that used the most recent observation as the forecast. Independent IMERG observations were used as a reference, and experiments were conducted to examine both overall statistics and case studies involving specific precipitation events. Overall, the forecasts provided by the *Nowcasting-Net* models are superior, with the *Convolutional Nowcasting Network with Residual Head* (CNC-R) achieving 25%, 28%, and 46% improvement in the test set MSE over the BM, LR, and RF approaches, respectively, for the Eastern CONUS. Results of further testing over the Western CONUS (which was not part of the training data) are encouraging, and indicate the ability of the proposed models to learn the dynamics of precipitation systems without having explicit access to motion vectors and other auxiliaries as features, and to then generalize to different hydro-geo-climatic conditions.

# 1- Introduction

[1] Precipitation has long been one of the most difficult aspects of weather to forecast. Just as ancient people were limited by environmental conditions when planning a hunt, modern people plan their everyday activities around cloudiness and the chance of rain. As weather patterns continue to be altered by climate change and the frequency of extreme weather events increases, it becomes ever more important to provide actionable predictions at fine enough spatiotemporal resolutions to be practically useful. Precipitation is a crucial driver for many natural hazardous events (e.g., landslides and flash floods), and its early prediction plays a crucial role in the development of new warning systems [1–8].

[2] The term "*nowcasting*" reflects the need for timely and accurate predictions of risky situations related to the development of severe meteorological events (World Meteorological Organization 2020). Such predictions facilitate effective planning, crisis management, and the reduction of loss to life and property. Most importantly, climate change has also recently led to more frequent catastrophic hydrometeorological events such as flash floods in various parts of the world, caused by more intense precipitation events [10]. As intense precipitation can damage life and property, accurate and reliable nowcasting of precipitation and other hydrometeorological forcings is extremely important [11, 12], and has emerged as a hot research topic in the hydrometeorology community. However, real-time, large-scale, and fine-grained precipitation nowcasting is a challenging task, due mainly to the inherent complexities of the dynamics of the atmosphere [13].

[3] Conventional methods for precipitation nowcasting include the use of storm-scale NWP models, radar echo extrapolation, and radar extrapolation coupled with NWP and/or stochastic field perturbations [14–37]. The radar-based forecasts are limited by ground clutter jamming, beam anomaly, signal attenuation, limited coverage over land, lack of coverage over the ocean, and installation costs. NWP methods have to deal with the computational expense of numerically solving partial differential equations subject to dynamic and thermodynamic laws, sensitivity to various kinds of noise, dependence upon initialization, the inability to exploit big data, and the prediction latency resulting from the numerical simulation and data assimilation steps. Accordingly, the development of reliable and timely nowcasting systems remains a critical need to prepare for natural hazards and to conduct related scientific investigations [38–41].

[4] Satellite remote sensing is the main observational source for near-real-time global precipitation fields with high spatiotemporal resolution. Satellite products rely mainly on precipitation estimates from infrared (IR) and passive microwave (PMW) observations. IR sensors aboard the geosynchronous-Earth-orbiting (GEO) satellites provide high spatiotemporal sampling, but precipitation estimates based on IR data are indirect and are mainly inferred from the cloud-top temperature. Meanwhile, PMW sensors offer more direct information related to precipitation, but are typically mounted only on the low-Earth-orbiting (LEO) satellites and, therefore, provide less frequent sampling [42, 43]. To use the strength of both IR and PMW information, merged precipitation products that combine IR- and PMW-based estimates have been developed such as the *Global Precipitation Measurement* (GPM) *Integrated Multi-satellitE Retrievals for GPM* (IMERG). However, the latency associated with these near-real-time global precipitation products is a few hours after the time of observation, which limits their application for many real-time tasks such as flash-flood and landslide warning systems that require timely observation [44, 45]. It is (in principle) possible to reduce this latency and improve their applicability, value, and impact by applying a robust nowcasting system to these products.

[5] Here we discuss the recent advances in deep learning (DL) that can help advance precipitation nowcasting. Increased computational power, and the emergence of novel DL structures and techniques, has led to breakthrough applications of deep neural networks (DNN) to areas such as autonomous vehicles, medical image analysis, object detection and tracking, semantic segmentation of images and point clouds, speech recognition, and natural language processing [46–49], among many others. One area of computer vision that has attracted much attention is the analysis of spatiotemporal data [50]. For example, [51] proposed a model for video frame prediction using spatially displaced convolutions, and [52] used a two-

phase algorithm for the same purpose by first predicting the flow in the video and then generating the desired frame using the predicted flow information. In this context, note that precipitation nowcasting is essentially a spatiotemporal sequence forecasting problem, where the sequence of recent precipitation maps serves as the input and the consequent sequence of a fixed number of future maps is the desired output. By examining recent advances in DL, with particular attention to recurrent neural networks (RNN), and more specifically long short-term memory (LSTM) models and convolutional neural networks (CNN), we can obtain useful insights into how this problem can be addressed.

[6] Machine learning (ML) and DL techniques have already been successfully applied in many areas of Geoscience such as precipitation retrieval [42, 43], time-series gap-filling [53], rainfall-runoff modeling [54, 55], and hydraulic conductivity estimation [56]. However, only a few studies have investigated precipitation nowcasting through DL [35–37, 45, 46, 56–62]. As an example, [59] showed that a *Convolutional LSTM* (ConvLSTM) model using radar maps outperformed the *Real-Time Optical Flow by Variational Methods for Echoes of Radar* (ROVER) method for precipitation nowcasting when tested over a very small region. While these studies demonstrated that DL-based methods outperform conventional methods, they all used radar images, focused on a very small region and trained the model using data from a specific type of precipitation storm (e.g., convective).

[7] In this study, we propose and test two DL-based precipitation nowcasting structures (hereafter referred to as *Nowcasting-Nets*) trained using IMERG precipitation fields. In theory, these structures should help address the dynamical spatiotemporal nature of the precipitation nowcasting problem. Our study area is the entire Contiguous United States (CONUS), and the data is representative of a variety of different kinds of precipitation events (frontal, convective, etc.), which enables us to assess the potential for real-world large-scale applicability of the approach. For comparison, we test two conventional ML methods, *Random Forest* (RF) and *Linear Regression* (LR), and also a persistence benchmark (BM) approach that uses the most recent observation as the forecast, to demonstrate the potential for improved accuracy and timeliness of precipitation nowcasting achievable using DL.

## 2- Dataset, Study Area, Methodology

### 2-1- Study Area

[8] The philosophy underlying DL suggests that if we propose a reasonable end-to-end representation for the model and have sufficient data to properly train it, we can obtain a good solution to the input-output mapping problem. For this study, we have chosen to focus on the Contiguous United States (CONUS) and to use sequences of precipitation maps over the Eastern CONUS to train and test the *Nowcasting-Nets* and other models. To further assess the capabilities of the models, the models were also tested over the Western CONUS, data from which were not used during the training and validation stages (*Figure S1*). In contrast to previous studies that used very local regions for model training and testing [61], our study includes data from a large area and therefore incorporates a variety of different types of precipitation events such as frontal, hurricane, and local convective systems.

### 2-2 Dataset

[9] Precipitation data were obtained from the *Global Precipitation Measurement (GPM) Integrated Multi-satellitE Retrievals for GPM* (IMERG) product [63], obtained from the Goddard Earth Sciences Data and Information Services Center (GES DISC) at https://disc.gsfc.nasa.gov. IMERG provides three gridded products with different levels of timeliness: Early run, Late run, and Final run. The IMERG Early and Late runs are quasi-real-time products, released with a lag of 4 and 12 hours from real-time, respectively. The Final run is a research-level product that is bias-adjusted using the *Global Precipitation Climatology Centre* (GPCC) monthly rain gauge data product [64] that has a time latency of about 3-4 months. All of the IMERG products are available at half-hourly 0.1° x 0.1° resolution and go back to June 2000. For this study, six years (2015-2020) of IMERG V06B Early run products were used.

### 2-3 Methodology

[10] Deep neural networks have been extensively used in a variety of different fields. One of the novel aspects of recent deep neural network technologies is their ability to effectively detect and extract the input data features that are most relevant and beneficial for the task at hand. This relieves the user from the need to perform the cumbersome, time-consuming, and challenging task of identifying and extracting the input data features that have the best explanatory power. For the problem of spatiotemporal precipitation prediction, this task becomes even more important given that the problem of how to relate temporal and spatial features and patterns seen in the data to the dynamics of how precipitation fields evolve is not well understood. Our strategy, therefore, was to perform experiments using state-of-the-art neural network model structures and to then design new structures inspired by them to accommodate our domain-specific needs. In particular, we investigated the ConvLSTM [59] and UNET [46] architectures, described below, and proposed 5 new structures that are inspired by them. These 5 models, together with three other ML-based methods, enable us to assess and compare the performance of a total of 8 data-based methods for precipitation nowcasting. Note that previous studies have already demonstrated the superiority of the DL-based models over conventional methods for precipitation nowcasting [39, 61]. The significant contribution of this study is to explore two DL structures that have the ability to simultaneously handle spatial and temporal data. The 8 methods investigated in this study are described below.

#### 2-3-1 UNET

[11] UNET models have generally been used for semantic segmentation of 2D and 3D images [46]. This network architecture consists of an encoder that maps the input image into a lower dimension space and a decoder that takes a tensor from this space and upsamples it to generate an output mask of the same size as the input image. The underlying intuition is that the lower-dimensional encoding causes the model to learn a set of useful features that contains information relevant to generating the desired output. The encoder uses Convolution and Max Pooling layers to generate the lower-dimensional hidden space representation, and

the decoder uses Deconvolution (Transpose Convolution) to expand the hidden space and generate the output mask. Several skip or residual connections [65] connect the encoder blocks to the decoder ones; these connections simply concatenate the filters from one of the encoder layers to the inputs of one of the decoder layers. One benefit of such connections is to alleviate the '*vanishing gradient*' problem and stabilize network training. The following equations explain how the model works.

$$B^D_i = MaxPool(W^D_{i2} * W^D_{i1} * I^D_i) \quad \text{(Equation-1)}$$
$$B^U_i = W^{UT}_i \otimes W^U_{i2} * W^U_{i1} * I^D_i \oplus I^U_i \quad \text{(Equation-2)}$$

[12] The first equation corresponds to the downsampling operation where $B^D_i$ represents the output of the i[th] block, $W^D_{i1}$ and $W^D_{i2}$ represent the weights of the first and second convolution layer in the i[th] block, $I^D_i$ is the input of the i[th] block and $*$ is the convolution operation. Similarly, the second equation corresponds to the upsampling operation where $B^U_i$ represents the output of the i[th] block, $W^U_{i1}$ and $W^U_{i2}$ represent the weights of the first and second convolution layer in the i[th] block, $W^{UT}_i$ is the weight of the deconvolution layer, $I^U_i$ is the input of the i[th] block, $\otimes$ represents deconvolution, and $\oplus$ represents concatenation.

[13] Inspired by this structure, in three of our proposed models we implement a 3D network architecture that is designed to handle spatiotemporal data and to generate an output sequence of precipitation maps given an appropriate input sequence. *Figure S2* illustrates the UNET structure that is used for semantic segmentation.

### 2-3-1-a Convolutional Nowcasting-Net (CNC)

[14] Inspired by the UNET architecture, we designed one of our base models, which we refer to as the *Convolutional Nowcasting-Net* (CNC). In a manner similar to UNET, the CNC architecture maps the input to a lower-dimensional space, and then back to the higher-dimensional space of the output sequence. However, unlike UNET, this model takes spatiotemporal data as inputs; specifically, it takes a time sequence (with length *n*) of precipitation maps as input and produces another time-sequence (with length *k*) of precipitation maps as output. The encoder performs spatial downsampling of the input and learns a compact tensor representation of the input sequence. The decoder performs a spatial upsampling of the compact tensor representation using the ConvTranspose layer and adjusts the temporal dimensions (sequence length) by appropriate padding during the convolution layer operations. Similar to UNET, residual connections link the encoder parts of the network to the decoder parts.

[15] *Figure 1* shows the structure of this model in detail. Based on the structure of the network and the kernel, padding, and stride of the layers illustrated above, the intuition is that the encoder learns spatial features from the input sequence and the decoder learns a mixture of both temporal and spatial features. However, squishing the input into the lower-dimensional hidden space is thought to reduce the spatial resolution [65]. To overcome this potential limitation, we used the CNC base model as a backbone and propose two related architectures, that we call CNC-R and CNC-D.

### 2-3-1-b Convolutional Nowcasting-Net with Residual Head (CNC-R)

[16] This model uses the CNC architecture as a foundation, but instead of learning the actual spatiotemporal features required to generate the desired outputs, it learns the features that help generate the '*residual*' between each timestep of the output and the final timestep of the current input sequence. The residuals are then added to the output of the previous time step to calculate the required outputs. *Figure 2* illustrates the architecture of this network. To implicitly learn the residuals of the output with respect to the last timestep of the input, instead of altering the input and output data, an addition layer is added to the output of the CNC model, which enables the backpropagation algorithm to properly update the weights of the internal layers for learning the residuals.

*2-3-1-c Convolutional Nowcasting-Net with Dual Head (CNC-D)*

[17] By combining the CNC and CNC-R model architectures, we obtain a hybrid model that can generate predictions of both the actual outputs and the residuals with respect to the last timestep of the input. This model has two branches corresponding to each of the aforementioned model architectures and their outputs. The fork-like structure of this model includes a shared encoder for learning the spatial features and two branches for learning the spatiotemporal features required to generate the actual outputs and the residuals. The ultimate output of the model is computed as the average of the two intermediate outputs of the two branches. *Figure 3* illustrates the structure of this model.

## 2-3-2 ConvLSTM

[18] ConvLSTM is a specialized class of neural networks that is specifically designed to handle spatiotemporal data. It combines the architectures of *Convolutional Neural Networks* (CNN), which are designed to handle image (spatial) data, and *Long Short-Term Memory* (LSTM) networks, which are designed to handle temporal data and are a subclass of *Recurrent Neural Networks* (RNN). So, similar to the LSTM, which sequentially processes temporal data while maintaining a memory of important aspects of what has previously been given to the network, the ConvLSTM maintains a memory of important aspects of the previous inputs images and states and uses convolution as the internal operation instead of matrix multiplication. *Figure S3* demonstrates the structure of the ConvLSTM and the way it works. The following equations represent the flow of data within one block of the ConvLSTM model.

$$i^t = \sigma(W_{xi} * x^t + W_{ai} * a^{t-1} + W_{ci} \circ c^{t-1} + b_i) \quad \text{(Equation-3)}$$
$$f^t = \sigma(W_{xf} * x^t + W_{af} * a^{t-1} + W_{cf} \circ c^{t-1} + b_f) \quad \text{(Equation-4)}$$
$$c^t = \circ f^t \circ c^{t-1} + i^t \circ \tanh(W_{xc} * x^t + W_{ac} * a^{t-1} + b_c) \quad \text{(Equation-5)}$$
$$o^t = \sigma(W_{xo} * x^t + W_{ao} * a^{t-1} + W_{co} \circ c^t + b_o) \quad \text{(Equation-6)}$$
$$a^t = o^t \circ \tanh(c^t) \quad \text{(Equation-7)}$$

where $x^t$ is the input sequence at time step $t$, $W_{xx}$ is a weight matrix, $a^t$ is the hidden state at time step $t$, $*$ is the convolution operation, and ∘ is hadamard product. More details about the equations and the ConvLSTM structure can be found in [59].

*2-3-2-a Recurrent Nowcasting-Net (RNC)*

[19] This model uses the ConvLSTM as a foundation for dealing with spatiotemporal data and implements ConvLSTM blocks to enable precipitation nowcasting. Each ConvLSTM block (see section 2-3-2) receives a current time step of the input and the previous memory (state) and produces one time-step of output and an updated memory state that feeds into the next step. Using this basic block, we created new structures designed to learn precipitation patterns, by stacking and unrolling ConvLSTM on sequences of precipitation maps. *Figure 4* illustrates our proposed structure.

*2-3-2-b Recurrent Nowcasting-Net with Residual Head (RNC-R)*

[20] Similar to the CNC-R architecture that learns spatiotemporal features relevant to generating residuals of the output with respect to the last timestep of the input, it is possible to have an RNC model with a residual head (i.e., RNC-R). We used the same technique as in the CNC-R to force the model to learn the residuals. The architecture of this model is shown in *Figure 5*.

[21] A property of ConvLSTM models is their large need for memory, during training, to enable the internal convolution operations to be performed. This computational barrier poses limitations to the development of a dual branch model similar to the CNC-D structure for RNC.

### 2-3-3 Model Training

[22] The aforementioned models were each fed with 9 time-steps of the IMERG Early product and trained to predict the subsequent 3-time steps, resulting in a forecast with 1.5 hours lead time. To initialize the weights of the deep learning models the Xavier method [66] was used, whose basic idea is to keep the variance of the inputs and outputs consistent, to prevent all output values from tending to zero. The Adam optimizer was used to train the network weights, by seeking to minimize the total precipitation nowcasting loss. For batch normalization, the moving average decay was set to 0.9, and a small float of 1E-5 was added to the variance term to avoid division by zero. The learning rate was initialized to 1E-3 and the batch size for training was set to 8. These hyperparameter values were selected based on exhaustive testing of different values via a grid search. All experiments are conducted using a TensorFlow V2 library [67], on a system with two AMD EPYC 7642 48-core CPUs, 470 GB of RAM, and a single NVIDIA V100S GPU. Dataset used for training and testing the Nowcasting-Nets is freely available at https://github.com/ariyanzri/Nowcasting-Nets.

### 2-3-4 Models Serving as Benchmarks for Comparison

[23] As mentioned earlier, in addition to the DL structures described above, we developed two other ML-based models – *Random Forest* (RF), *Linear Regression* (LR), and the persistence benchmark (BM). The capacity of RF and LR to learn spatiotemporal features is very limited, and such methods are typically used for simple classification and simple trend prediction tasks. The RF and LR models were fed with the same data as the DL models, with the difference that they were fed pixel by pixel 9-step sequences of data, rather than the entire images, and were trained to output 3-step sequences at each pixel which were then assembled to create precipitation maps. The BM model simply repeats the last final time step of the input sequence as the prediction for the next three timesteps. While this might seem rather naïve, it is worth noting that given the relatively short forecast lead time (1.5 hours), the BM prediction can be robust and difficult to beat. The RF and LR models were developed using Sci-kit learn library [68].

# 3- Results and Discussions

[24] Several experiments were performed to assess the performance of the models and methods used for nowcasting. The assessment included analysis of overall statistics based on all of the events studied, and several case studies using a selection of individual events.

[25] The purpose of the models examined here is to provide short-term nowcasts of the dynamic spatiotemporal evolution of precipitation events. A performance evaluation based on metrics that measure grid-by-grid differences (e.g., mean squared error (MSE), correlation coefficient (CC), etc.) can be incomplete, because they may not provide a good assessment of the models' abilities to reproduce the shapes and organizational structures of the events. A more comprehensive evaluation should also include a visual assessment of the actual precipitation maps and their spatiotemporal evolution. We provide such an assessment through several case studies.

*Figure S4* shows nine consecutive ½ hourly precipitation maps from five different precipitation events (rows A-E) that fell into our study domain. The precipitation maps are from the IMERG Early product. As can be seen, the shapes and intensities of the precipitation systems change significantly through each sequence. These maps are fed as input to each of the trained models and the results are reported and discussed in the following sections. Storm events A and B will be used for assessing predictions with +90 minutes lead times, events C and D will be examined for extended prediction with +4.5 hour lead times, and event E is used for testing the models over the Western US, where no training was performed.

## 3-1 Prediction with +90 minutes Lead Time

### 3-1-1 Overall Performance of Models on the Testing Data Set

[26] We first checked the performance of the models on an independent testing set that includes 533 precipitation events (from both summer and winter), each with an input sequence of 9 time-steps and the output consisting of 3 time-steps. Two types of metrics were used – continuous (see *Figure 6*) and categorical (see *Figure 7*). The continuous metrics include the Mean Squared Error (MSE), bias ratio (BIAS), $R^2$, and the correlation coefficient (CC); see definitions provided in Appendix A.

[27] The CNC-D and CNC-R architectures achieved the best MSE performance (*Figure 6-A*), followed by CNC for which the three lead time forecasts have mean MSEs values of 0.762, 0.762, and 0.942, respectively. Importantly, CNC model MSE performance (optimal value is 0.0) does not decline with increasing lead time and the corresponding mean performance declines for CNC-D and CNC-R are relatively small. In contrast, MSE performance of the baseline persistence benchmark (BM) scenario declines significantly with increasing lead time (from 0.891 to 1.442 to 1.905), which is expected given that BM is very sensitive to rapid changes in the deformation, intensity, and location of the precipitating system. Perhaps surprisingly, the performance of the *Random Forest* (RF) and *Linear Regression* (LR) models is fairly good. It can be seen that the RNC models have relatively large errors, with RNC-R performance being similar to that of BM. Overall, the MSE performance of the CNC-D and CNC-R models is approximately 46%, 28%, and 25% better than BM, LR, and RF models, arising from the ability of the CNC architectures to deal with spatiotemporal sequences. In general, the CNC architecture exhibits superior MSE performance than the RNC architecture.

[28] In terms of BIAS (*Figure 6-B*; optimal value is 1.0, values larger/smaller than 1.0 indicate overestimation/underestimation), the BM, LR, and RF models scenarios have the best performance (mean BIAS is 0.975, 0.989, and 0.986, respectively). The RNC model exhibits significant overestimation (mean BIAS is 1.972, 2.002, and 1.991 for the first, second, and third lead time forecasts respectively) while the inclusion of the Residual Head (RNC-R) results in improved bias (mean BIAS is 0.952, 0.894, and 0.825 for the three lead times). This indicates that the addition of skip connections to the RNC model helps reduce its predictive bias. However, the underestimation bias of the RNC-R predictions does get progressively larger with increasing lead time. In contrast, the overestimation bias of the CNC model (~ 1.25) remains

almost constant as the lead time increases, indicating that its performance is stable over time. CNC-D underestimates slightly while CNC-R overestimates.

[29] In terms of $R^2$ (*Figure 6-C*; optimal value is 1.0) and CC (*Figure 6-D*; optimal value is 1.0), the CNC-D and CNC-R architectures provide the best performance (forecast lead time $R^2$ values are approximately 0.637, 0.515, and 0.396 for CNC-D and 0.665, 0.514, and 0.377 for CNC-R). In comparison with the RF, LR, and BM forecasts, the CNC-D and CNC-R models show 40%, 50%, and 1440% improvement. Further, CNC model performance, despite being weaker than that of the CNCs with skip connections, remains stable for all lead times. The results reveal that, while the addition of the most recent time step as the skip connection improves model performance in general, the performance declines as lead time increases. This may cause serious issues for longer lead times, which will be explored later in this paper. *Figure 6-C* shows that RNC architecture performance is weaker than that of CNC (mean $R^2$ is 0.248 compared to 0.436). RNC-R and the persistence BM method have the weakest performance among all models.

[30] Overall, these results indicate that: (1) The CNC and RNC architectures achieve the most stable results, with performance that does not decline over the forecast lead times tested, and with RNC exhibiting generally weaker performance; (2) The CNC-R architecture provides the best performance in terms of all metrics except bias; (3) Performance of the persistence benchmark BM approach drops significantly with increasing lead time; and (4) The relatively simple RF and LR models achieve surprisingly good performance. The latter finding is interesting, considering that these methods are not specifically designed to represent the temporal dynamics of precipitating systems; we will explore this issue further in the next set of experiments.

[31] *Figure 7* compares model performance in terms of four categorical metrics – the probability of detection (POD), false alarm ratio (FAR), Heidke skill score (HSS), and accuracy score (ACC); see definitions provided in Appendix A. In terms of POD (*Figure 7-A*; optimal value is 1.0), the CNC-D architecture achieves the best performance (mean PODs for the three lead times are 0.828, 0.799, 0.795, respectively), followed by CNC-R, and RF. POD performance for the persistence BM approach decreases with increasing forecast lead time (0.777, 0.698, and 0.659, respectively). Again, performance of the CNC architecture is stable (~ 0.757), and slightly better than RNC (0.669) and RNC-R (0.756). On average (over +30, +60, and +90 minute lead times), CNC-D and CNC-R architectures have 16% and 8% better performance than the persistence benchmark, 18% and 9% better performance than LR, and 3% and 2% better performance than RF.

[32] In terms of FAR (*Figure 7-B*; optimal value is 0.0), the CNC-R architecture and the persistence BM approach achieve the best performance (mean FAR values are 0.271 and 0.275, respectively). Notably, the DL and ML models tend to produce very small precipitation rate values (e.g., 1E-5), which affects their performance in terms of categorical metrics, but this problem does not plague the persistence benchmark BM approach which simply repeats the IMERG estimate.

**[33]** In terms of the HSS (*Figure 7-C*; optimal value is 1.0), which can be considered to be an 'overall' score, the LR and RF methods are characterized by sharp declines with increasing lead time. Again, the CNC architectures achieve the best performance (mean HSS is 0.707 for CNC-R, 0.684 for CNC-D, and 0.671 for CNC). Finally, almost all of the models outperform the benchmark scenario in terms of accuracy score (ACC).

### 3-1-2: Case Studies and Visual Assessment

[34] The model-based predictions for event A are shown in *Figure 8*, where each row shows the forecast for a different lead-time – for +30, +60, and +90 minutes, respectively – while each column shows the prediction provided by a specific model. Each predicted image is accompanied by the corresponding MSE, BIAS, and CC performance statistics. The final column shows the actual (observed) IMERG image that is used as a basis for comparison.

[35] The results show that the basic RNC architecture strongly underestimates precipitation intensities (BIAS is ~ 0.5), while the RNC-R architecture with the residual connection reduces this bias but is unable to reproduce the temporal development of the system, as indicated by the decreasing CC and increasing MSE and BIAS with increasing lead time. Overall, the CNC-R and CNC-D architectures exhibit the best performance amongst all of the models, tracking the shape and intensity changes of the system well and achieving the highest CC and lowest MSE. The basic CNC architecture does also track the shape of the system well but tends to underestimate the higher precipitation intensities. Meanwhile, the simpler LR and RF models predict the first-time step with acceptable accuracy (BIAS is almost perfect and the CC is ~ 0.7), but their performance deteriorates significantly with increasing lead time.

[36] As expected, due to rapid change in the shape and intensity of the system, the performance of the BM persistence benchmark declines significantly with increasing forecast lead time (e.g., CC drops 16% from +30 min to +90 min lead time). In contrast, the CNC-R and CNC-D architectures provide ~ 35% better forecasts in terms of CC and ~60% better forecasts in terms of MSE compared to the persistence benchmark. Overall, *Figure 8* indicates that the models based on the CNC architecture are superior to the others at predicting the dynamic spatiotemporal development of the system.

[37] The model-based predictions for event B (see the second row of *Figure S4*) are shown in *Figure 9*, which corresponds to the spatiotemporal development of a tropical storm system (Hurricane Fred) over the ocean east of the Florida Peninsula. The input sequence indicates that the system is strengthening with time. As with event A, the RNC architecture underestimates the storm intensities (BIAS is ~ 0.4 at all time steps) while weakly capturing its spatiotemporal development (CC is ~ 0.66 at all time steps). The addition of skip connections improves the ability of RNC-R to predict the intensity of the system but it progressively overestimates as the lead time increases (BIAS is 1.02, 1.06, and 1.17 for the three forecast lead times). It appears that the skip connections interfere with the internal state of the ConvLSTM cells because the quality of the RNC-R predictions is similar to that of the BM persistence benchmark. The CNC-D and CNC-R architectures both predict the development of the system well, while the CNC without skip connections has weaker performance (BIAS is ~ 0.7). The simpler LR and RF models underestimate the system precipitation intensities while showing skill in capturing the development of its shape/structure (CC is 0.87, 0.78, and 0.67 for both models). As expected, the skill of the BM persistence benchmark is poor due to the rapid development of the system (MSE increases from 1.12 to 3.03 mm/hour). Overall, the CNC-based models perform relatively well, compared to the others.

## 3-2 Extending the Nowcast Lead Time to +4.5 Hours by Feeding the Models with Their Predictions (Feedback Loop)

### 3-2-1 Overall Performance of Models on the Testing Data Set

[38] The IMERG early product is available with about 4 hours delay past real-time to allow the inclusion of all or most of the input satellite precipitation products into it. For several applications, such as hazard warnings (e.g., for flood and landslide), the 4-hour latency is still significant and can negatively impact its usefulness. Here, we investigate the skill of each model to predict precipitation with a +4.5-hour lead time, so that if applied to IMERG Early it would enable a user to access "*real-time*" estimates of precipitation with no latency. To achieve this, each model was first fed with the input sequence, and then its predictions were recursively fed back into it as input until the lead-time of 4.5-hours is reached. This requires running each model three times, each time producing a nowcast with a 1.5-hour lead time. This experiment helps to assess the range of the model's capabilities and to check whether the models can reproduce the dynamical evolution of the precipitation systems.

[39] We first examine the performance of the models over the entire testing data set, and then investigate two selected case studies (Figure S4; events C and D). *Figure 10* shows how the mean value of model performance (in terms of four continuous metrics) varies with increasing lead time in ½ hour increments. *Figure 10-A* shows that while CNC-R and CNC-D have similar MSE performance at the first three time steps, which was also observed in *Figure 7*, as the lead-time increases the CNC-R architecture achieves better performance (less rapid decline) than CNC-D. This suggests that the extrapolation performance of CNC-D is affected by the head, similar to the CNC model. The performance of RNC-R is no better than that of the BM persistence benchmark, indicating that adding skip connection significantly affects its performance. The RF and LR models achieve similar performance to each other, as was also observed in previous experiments.

[40] In terms of BIAS, the RNC architecture provides the worst performance, while the simple RF and LR models give the best performance. The CNC-R architecture, although performing relatively well for the first few time steps, progressively overestimates as the lead time increases. This might be due to adding the last time step of the input sequence as the skip connection to the predictions, which is not very beneficial as the lead time increases considering the fast development in the intensity and deformation of the systems (as observed in the input sequences shown in *Figure S4*).

[41] In terms of $R^2$, the CNC-R and CNC-D architectures have similar performance for the first three time steps, while CNC-R performs better than CNC-D at longer lead times and is significantly better after the 7$^{th}$ time step, suggesting that CNC-R may better reproduce the spatiotemporal dynamics of precipitating systems. As before, the performance of RNC-R and BM drop sharply as the lead-time increases. The addition of the skip connections in the RNC-R probably interferes with the internal state of the ConvLSTM which is the core of the RNC model. The CNC-R architecture has the best performance in terms of CC. Overall, the CNC-R architecture outperforms the other models at predicting the dynamical spatiotemporal evolution of the precipitation systems with increasing lead times.

### 3-2-2: Case Studies and Visual Assessment

[42] Next, we examine feedback loop model performance for two case studies (i.e., a frontal and a convective system). *Figure 11* shows the results for a frontal system event on August 15, 2015 (see the third row of *Figure S4* for the input sequence of this system). Each row shows a model, and each column represents a time step starting at 18:00 UTC. From the input sequence and the observations (last column of *Figure 11*), we can see that the precipitation system moved, deformed, and presented different intensities over time.

[43] The RNC architecture largely underestimates the precipitation intensities and is unable to capture the development of the system. The performance of RNC-R is also very poor, due to the skip connections as seen in previous experiments. The performance of CNC-D is better than for the RNC models; however, with increasing forecast lead time a noisy response emerged, clearly seen as wide-spread light rainfall. This behavior is probably due to the CNC head because similar behavior is also observed for CNC. The CNC-R architecture seems to mitigate this issue, resulting in the best performance among all models, although a tendency to underestimate persists with increasing forecast lead time. The CNC-R and RF are best at capturing the change in the shape of the system, but RF produces a large amount of noise (seen as extensive light intensity precipitation). CNC exhibits acceptable performance for the first few time steps, but its predictions become progressively noisier at later time steps. Not surprisingly, the performance of LR is not very good, considering that it can capture either a decrease or increase in precipitation rate but not both. By design, the BM persistence benchmark does not track the changes in the shape and intensity of the system over time.

[44] *Figure 12* corresponds to a convective system that decays, changes shape, and moves to the west over time. The input sequence is shown by the fourth row of *Figure S4*. The RNC architecture correctly predicts that the system is getting weaker, but largely underestimates the intensity of the precipitation. RNC-R overestimates intensities, probably due to the skip connections. CNC-D tracks the system behavior well for the first few time steps but progressively underestimates with increasing lead time. CNC-R achieves the best performance, tracking the change in the intensity, shape, and movement of the system relatively well. However, by moving towards a longer lead time, the predicted maps become smooth and lose details. CNC has similar performance to CNC-D but underestimates the system even more. The simpler LR and RF models capture the fact that the system is getting weaker over time, but both tend to underestimate. Of course, BM overestimates significantly with increasing lead time and does not track the decay of the precipitation intensity over time. Overall, *Figure 9* shows that forecasting precipitation events beyond a few hours is very challenging for all of the models tested here, which is perhaps not surprising given that they are trained to optimize forecasting performance over three timesteps only.

### 3-3 Testing for Storms in the Western US

### 3-3-1 Overall Performance of Models on the Testing Data Set

[45] All of the models were trained with precipitation maps from the Eastern CONUS. In this section, we investigate how well they perform for precipitation systems over the Western CONUS, to assess the extent to which the relationships learned by the models can be applied over other regions. In other words, we assess whether the dynamical precipitation patterns learned for the Eastern CONUS can be used to predict the dynamics of precipitation systems over the Western part of the country that has a different hydro-geo-climate. First, we examine overall model performance on an independent test set composed of 533 events over the Western CONUS using continuous metrics. A case study is then examined for visual assessment.

[46] *Figure 13* indicates that the CNC-R, RNC, and RF models exhibit consistent MSE performance with increasing lead time. In contrast, the other models perform well only for the first few time steps and lose skill thereafter. CNC and CNC-D exhibit the worst performance. RNC-R and CNC-R have better BIAS performance than the other methods, while the remaining models either overestimate or underestimate significantly as the lead time increases. RNC and LR show larger BIAS than all of the other models. CNC-R and RNC have the best $R^2$ performance and outperform other models (especially CNC-D, CNC, and BM). CNC-R, the RNCs, and BM have the best CC performance. The summary of metrics indicates that the CNC-R architecture achieves the best overall performance, generally consistent with the previous results. However, by comparing the CNC-R prediction with +4.5 hours lead time with a prediction with +30 minutes lead time, a 77% reduction in CC and 130% increase in MSE can be seen.

### 3-3-2: Case Studies and Visual Assessment

[47] A precipitation event in the Western U.S. is used to visually compare the performance of the method studied for precipitation prediction, with lead time up to +4.5 hours, through the repeated feedback loop experiment (*Figure 14*). The input sequence for this case study is presented in the fifth row of *Figure S4*. This storm features large changes in intensity and shape and moves towards the Southeast. As before, RNC underestimates the storm intensities significantly while RNC-R overestimates. The CNC-D architecture captures the decay of the system but adds a significant amount of light rainfall as time progresses. CNC-R outperforms other methods in reproducing the dynamics of precipitation intensity but is unable to accurately track the location of the system, similar to the other models. The performance of CNC is similar to that of CNC-D. The simpler LR and RF models exhibit very little skill in tracking the development of the system as the lead time increases. The BM persistence benchmark performs poorly due to its inherent inability to track variations in both intensity and location of the system.

[48] Overall, the predictions are weaker in the Western CONUS, which is expected given that the models were trained with data from the Eastern CONUS. However, it is interesting that the CNC-R architecture shows some skill in capturing the movement patterns and intensity dynamics of the system, suggesting that the model can learn some common features related to precipitation dynamics between the Eastern and Western CONUS.

## 4- Concluding Remarks

[49] Precipitation nowcasting is critical for the successful implementation of hazard warning systems, such as for flooding and landslides. We developed and tested two state-of-the-art DL structures for precipitation nowcasting, including 5 different model architectures, and compared them against classical ML techniques (RF and LR) and a persistence benchmark (BM). The proposed *Recurrent Nowcasting-Net* (RNC) architecture utilizes *ConvLSTM* cells, while the *Convolutional Nowcasting-Net* (CNC) architecture is based on the UNET structure. Five variations with and without skip connections were developed and, together with RF, LR, and BM, a total of 8 nowcasting methods were compared against observations for forecast lead times up to 1.5 hours. By using a recursive feedback loop approach, the capability of the model to extend the forecasts up to +4.5 hours was also investigated.

[50] Our results indicate that models based on the CNC architecture provide better results than those based on the RNC architecture. Overall the best performance is achieved by the CNC-R architecture, which uses a 3D convolutional neural network that learns which spatiotemporal features are useful for predicting the residual between the last timestep of the input sequence and each timestep of the output. This suggests that learning the residuals (differences in maps between time steps) is a better strategy for the dynamical spatiotemporal prediction of precipitation events. In general, the proposed architectures were able to learn useful features relevant to the evolution of precipitation patterns in time and space. Note that our approach used only the information provided directly by precipitation maps for nowcasting, and other potentially useful auxiliary data such as motion vectors were not used here.

[51] The models based on the CNC architecture, and in particular CNC-R, showed good overall predictive skill in terms of categorical (POD, FAR, HSS, and ACC) and continuous (MSE, CC, R2, and BIAS) skill score metrics on all of the events in the independent test set (including 533 events [convective, frontal, etc.]) and individual case studies. To recap, the CNC-R model provided 46%, 28%, and 25% better BIAS performance than the BM persistence benchmark, and the simpler LR, and RF models respectively. Similarly, the CNC-R model shows 1440%, 40%, and 50% better performance in terms of $R^2$, and 16%, 18%, and 3% better performance in terms of POD compared to BM, LR, and RF, respectively. These gains in performance arise from the ability of the CNC-R architecture to learn from spatiotemporal data and to extract prediction-relevant features from sequences of precipitation maps.

[52] For the 5 case studies investigated, the CNC-R architecture consistently demonstrated superior performance over the other models, in terms of skill scores (60% decrease in MSE and 35% improvement in CC over the persistence benchmark), and in the ability to track the changes in shape and intensity of the precipitation systems over time. CNC-R also showed reasonably good performance at extending the forecasts up to +4.5 hours, when implemented in recursive feedback mode, for the Eastern and Western CONUS, indicating its ability to learn the spatiotemporal dynamics of weather systems evolution.

[53] While this and previous studies have demonstrated the superiority of DL-based precipitation nowcasting over conventional nowcasting approaches (such as NWP-based and radar echo extrapolation), some challenges need specific attention and should be addressed in future studies. First, given that the use of DL for precipitation nowcasting is still in its early stage, it is not yet clear how models should be evaluated to take into account the needs of real-world applications. Previous studies mainly used precipitation images from a single radar observatory that covers small regions. Here, the study area includes the entire CONUS, and our experiments were designed to test the model at this scale so that it is likely to be more relevant for real-world applications.

[54] Second, for effective nowcasting, high-dimensional spatiotemporal sequences with multi-step predictions must be made. Because effective training of deep neural networks requires massive amounts of data and computation power, the computational demands can limit testing various approaches. For example, we were unable to train and test a *Recurrent Nowcasting-Net with Dual Head* (RNC-D) architecture due to the exorbitant computational expense. While the enhanced skill of DL models comes with additional

computational costs that can inhibit their development, when trained, they are often quick and efficient to implement, providing a clear advantage over conventional methods.

[55] Third, the rapid changes in intensity, shape, and direction of precipitation systems can make the prediction task difficult. Arguably, the problem of precipitation nowcasting is considerably more challenging than comparable tasks in the computer science field such as video prediction. This is mainly because the combined effects of the changes in intensity, deformation, and movement of objects/fields is not common in video prediction. Further, the chaotic characteristics of atmospheric systems tend to result in declining predictive skills with increasing forecast lead time. Of course, this latter issue is not specific to DL-based models and physics-based approaches must also contend with the same issue.

[56] Fourth, the problem of blurriness of images that is widely reported in the DL literature remains unresolved. In this study, we have taken advantage of skip connection and the use of several blocks with batch normalization to help diminish this issue.

[57] Last but not the least, DL and ML models both require good quality training data sets. Noisy and erroneous observations can adversely affect the ability of the models to extract information regarding the processes that control the spatiotemporal evolution of precipitation patterns. With more powerful computational resources and more accurate and informative sources of information (e.g., from remote sensing satellite platforms), a bright future for the nowcasting of precipitation using DL methods can be expected.

# Acknowledgments

Financial support was made available from NASA MEaSUREs (NNH17ZDA001N-MEASURES) and NASA Weather and Atmospheric Dynamics (NNH19ZDA001N-ATDM) grants. The dataset used in this study and the *Nowcasting-Nets* are freely accessible for non-commercial use at https://github.com/ariyanzri/Nowcasting-Nets accessed on 08 August 2021. We also acknowledge the University of Arizona Data Science Institute (https://datascience.arizona.edu/ accessed on 8 August 2021) and CyVerse (https://cyverse.org/ accessed on 8 August 2021) especially Dr. Eric Lyons, Nirav Merchant, and Maliaca Oxnam who have provided moral support and access to high-performance computing resources.

# Appendix I: Definitions of the Statistical Metrics

Quantitative statistics are obtained using estimated (est) and observed (obs) quantities. Categorical statistics are obtained using the contingency table. The quantitative and categorical statistics used in the present work are calculated as indicated below, where N is the total number of observed and estimated precipitation pairs, and H, F, M, and Z are the numbers of hits, false alarms, misses, and correct negatives respectively.

$$POD = \frac{H}{H + M}$$

$$FAR = \frac{F}{H + F}$$

$$HSS = \frac{2(HZ - FM)}{(H + M)(M + Z) + (H + F)(F + Z)}$$

$$ACC = \frac{H + Z}{N}$$

$$CC = \frac{\sum_{i=1}^{N}(obs_i)(est_i) - N(\overline{obs})(\overline{est})}{\sqrt{[\sum_{i=1}^{N}[(obs)^2 - N(\overline{obs})^2]\sum_{i=1}^{N}[(est)^2 - N(\overline{est})^2]}}$$

$$BIAS = \frac{\sum_{i=1}^{N}(obs_i - est_i)}{\sum_{i=1}^{N} obs_i}$$

$$MSE = \frac{\sum_{i=1}^{N}(obs_i - est_i)^2}{N}$$

$$R^2 = 1 - \frac{\sum_{i=1}^{N}(obs_i - est_i)^2}{\sum_{i=1}^{N}(obs_i - \overline{est})^2}$$

# Figures

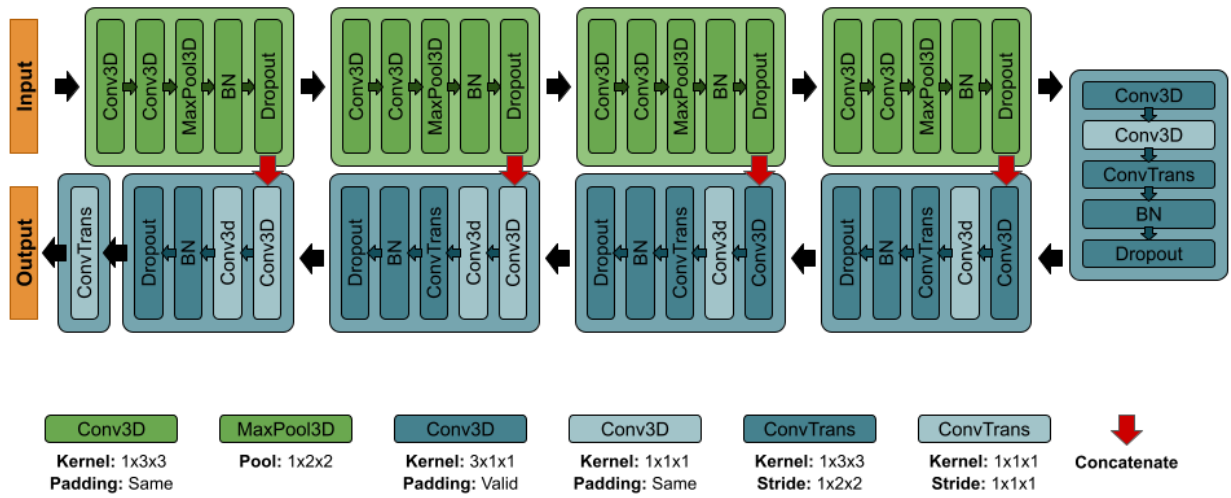

*Figure 1:* General architecture of the *Convolutional Nowcasting-Net* (CNC) model. Note that BN indicates the Batch Normalization Layer which is thought to increase the generalization ability of the model and make the training process more stable.

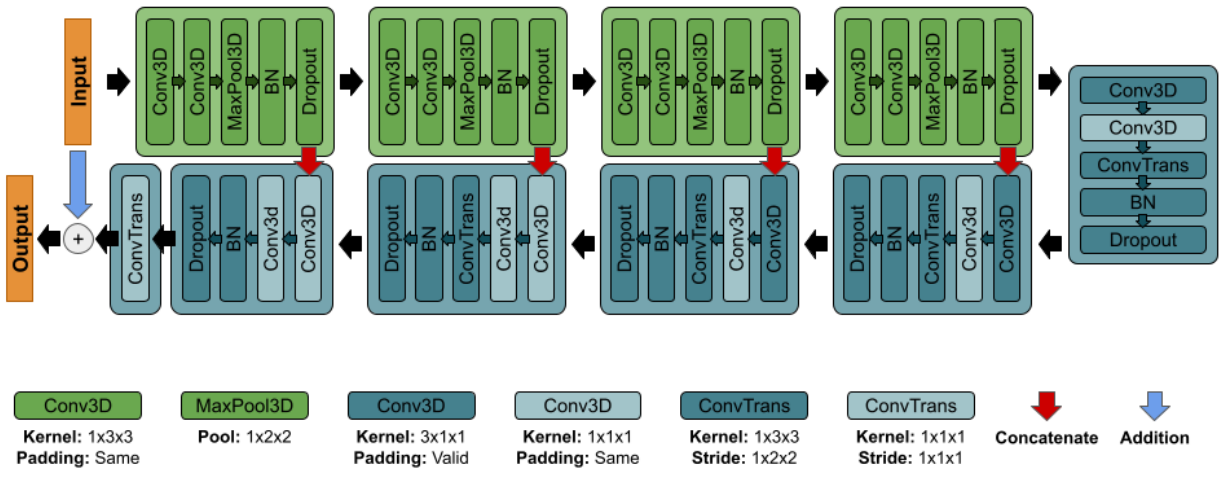

*Figure 2:* General architecture of the *Convolutional Nowcasting-Net with Residual Head* (CNC-R) model.

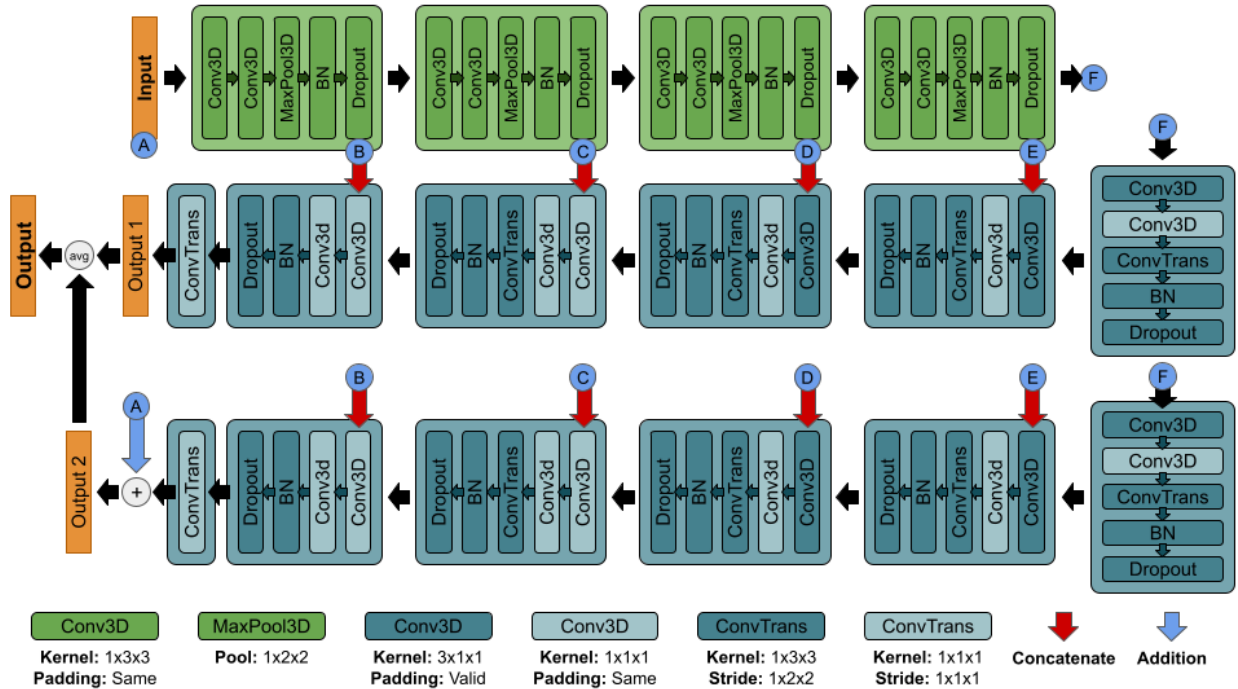

*Figure 3:* General architecture of the *Convolutional Nowcasting-Net with Dual Head* (CNC-D) model.

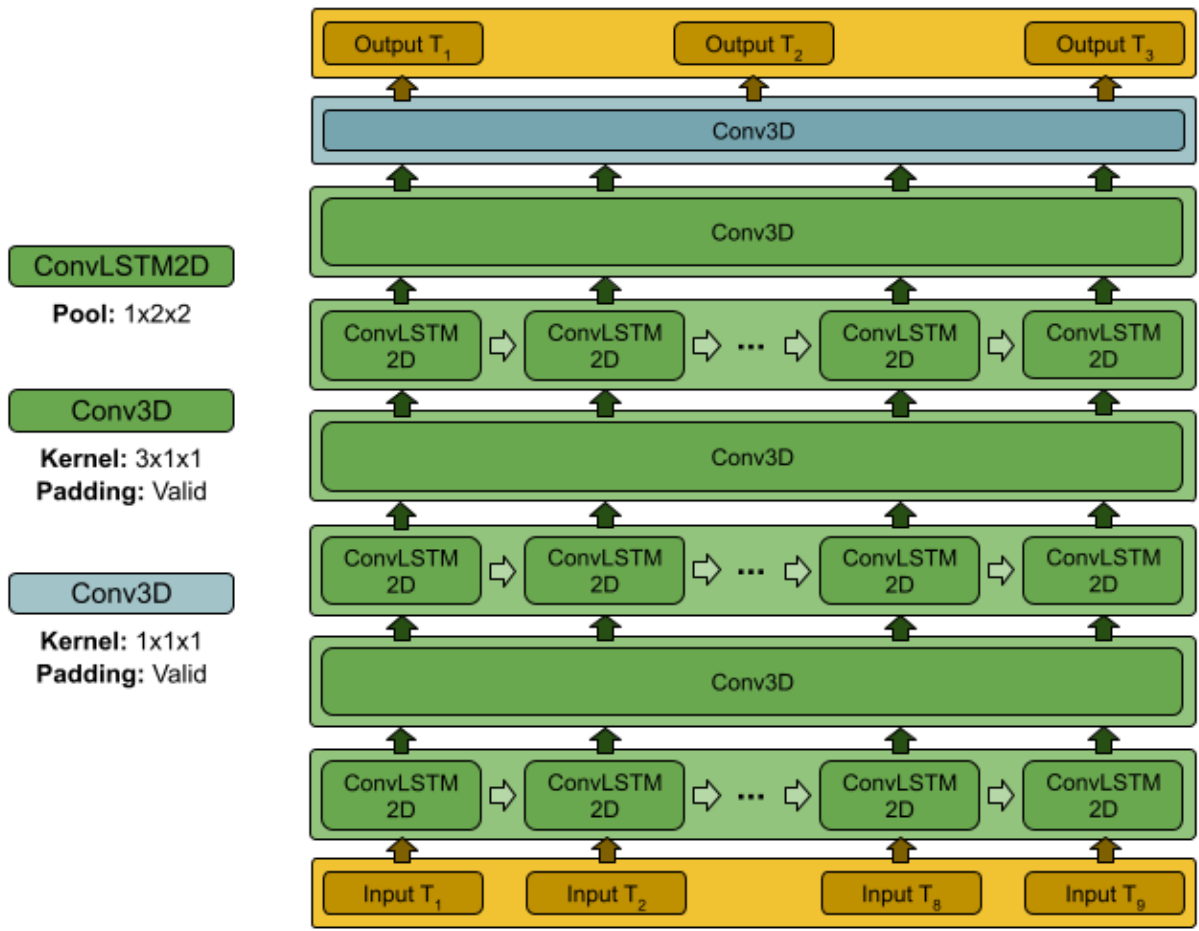

*Figure 4*: General architecture of the *Recurrent Nowcasting-Net* (RNC) model.

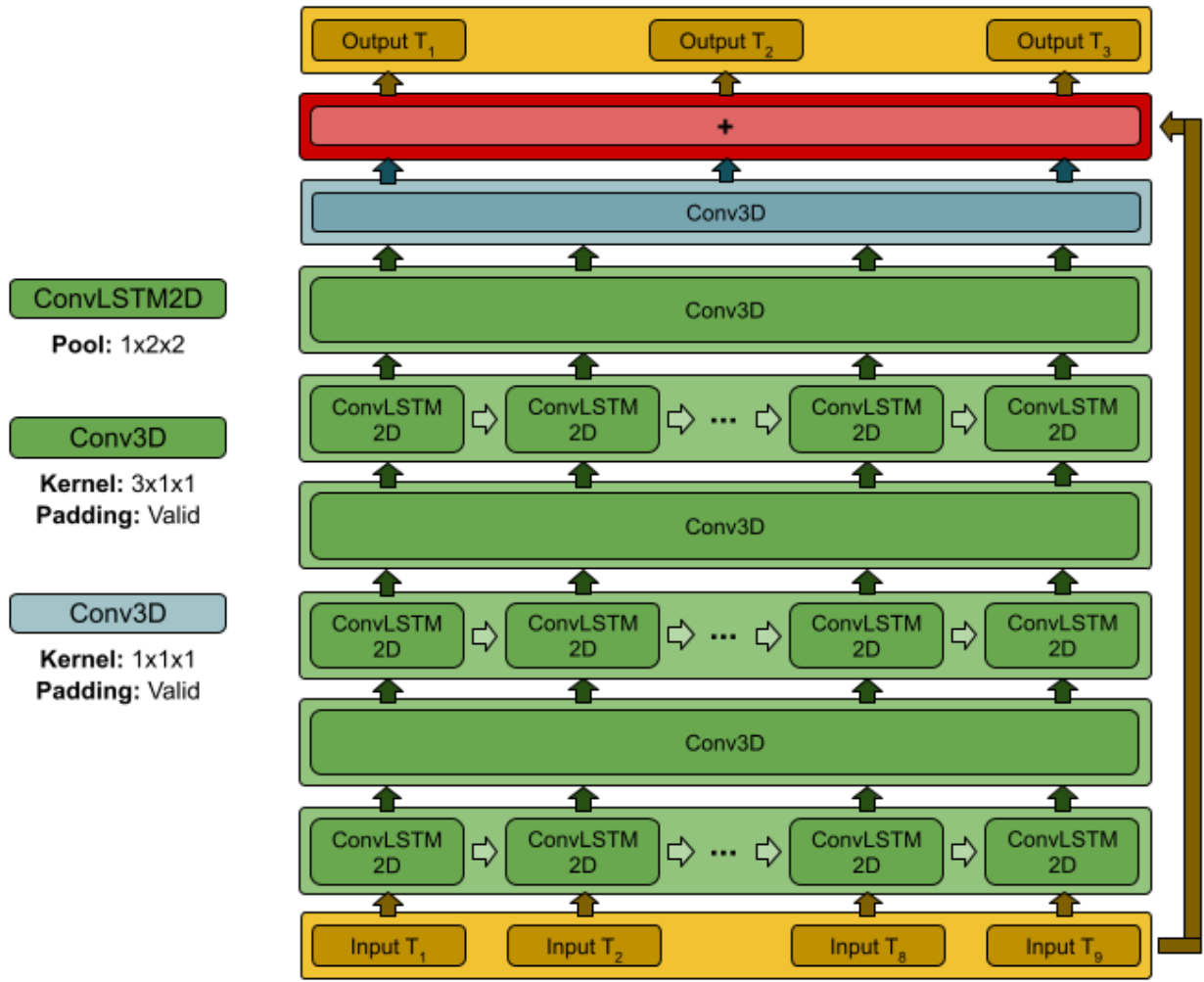

*Figure 5*: General architecture of the *Recurrent Nowcasting-Net with Residual Head* (RNC-R) model.

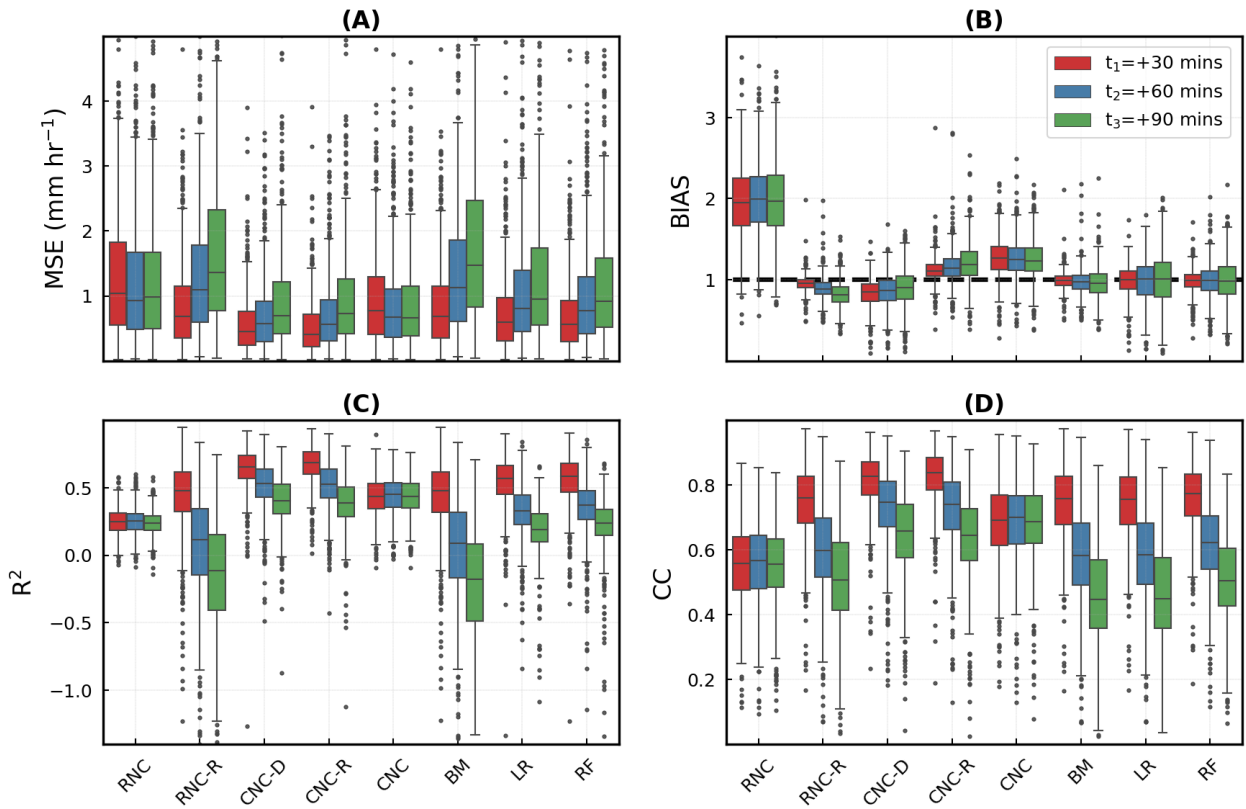

*Figure 6:* Boxplots of continuous metrics for models developed in this study. Panel (A): Mean Squared Error (MSE), Panel (B): BIAS, Panel (C): $R^2$, and Panel (D): correlation coefficient (CC). Red, blue, and green colors correspond to lead times of +30, +60, and +90 minutes respectively.

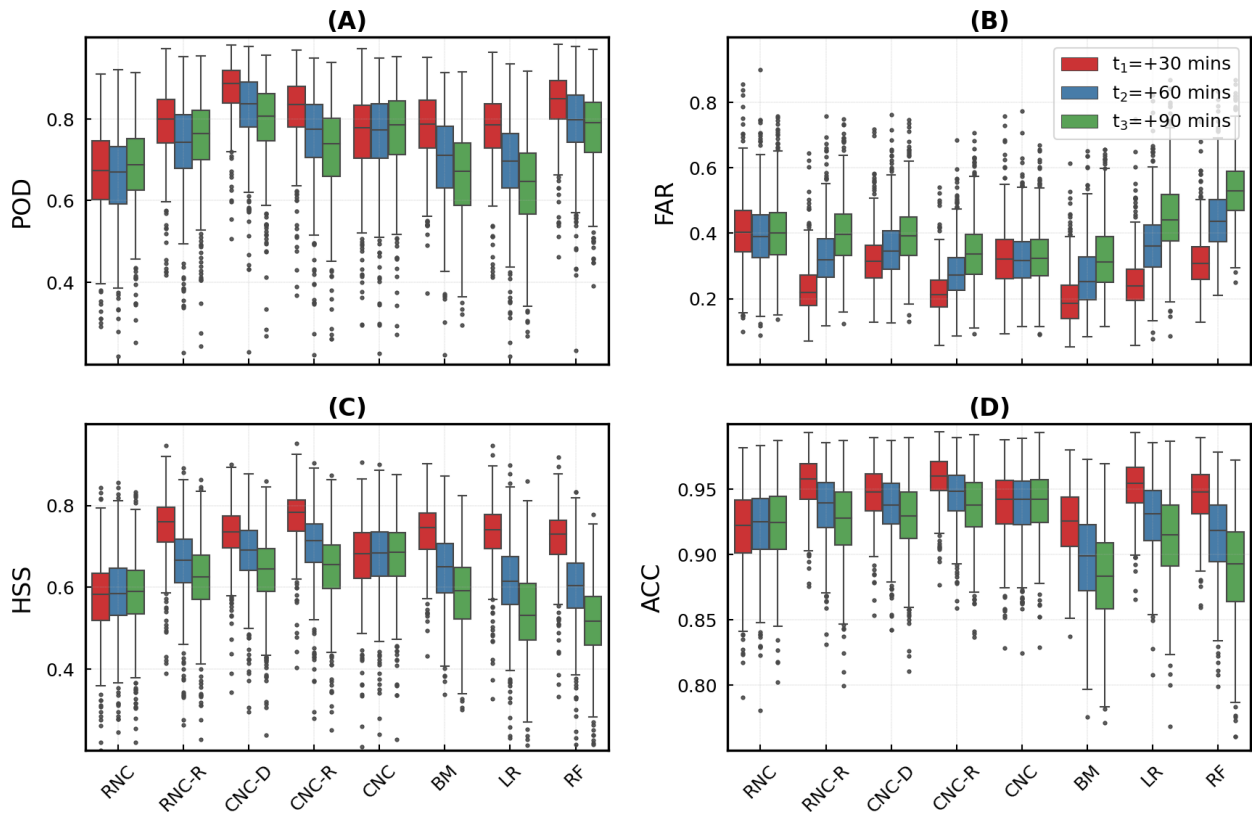

*Figure 7:* Boxplot of the categorical metrics for different models developed in this study. Panel (A): the probability of detection (POD), Panel (B): false alarm ratio (FAR), Panel (C): Heidke skill

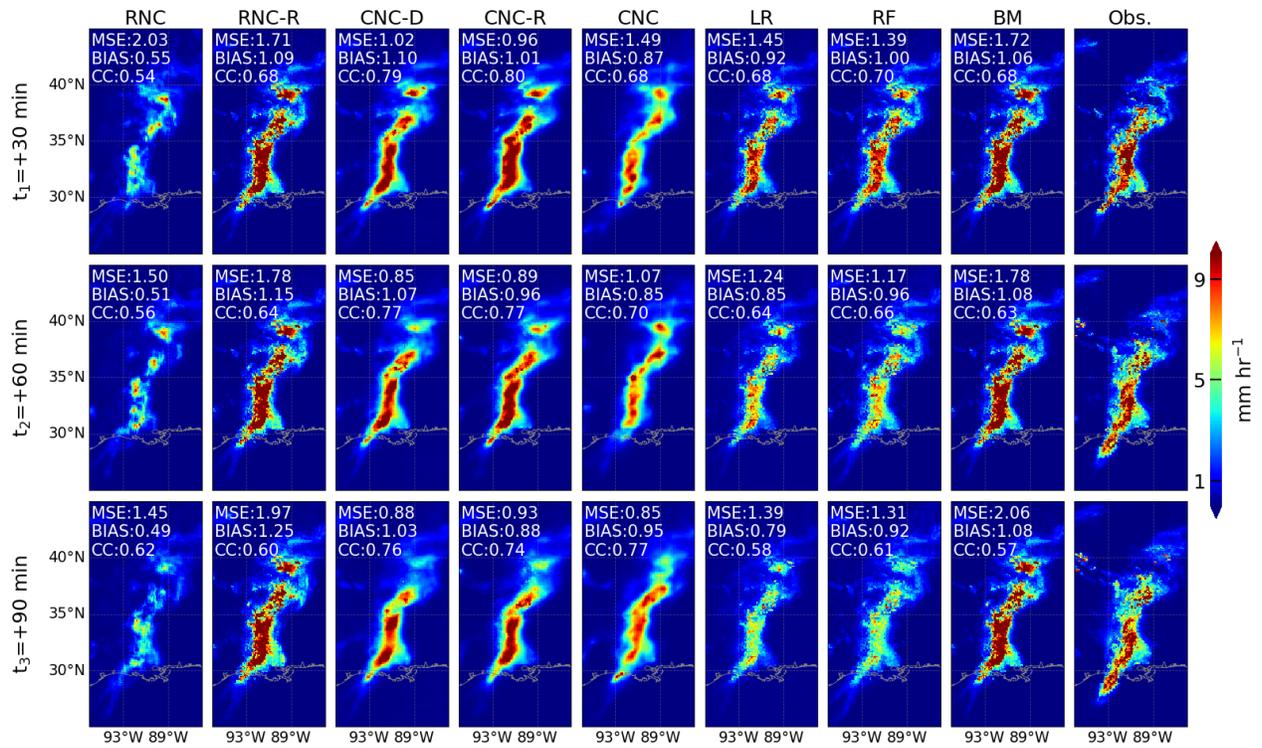

*Figure 8:* Precipitation nowcasting for storm A on July 18, 2015, using different models developed in this study up to 1.5 hours lead time. Each row shows a time step, and each column represents a model.

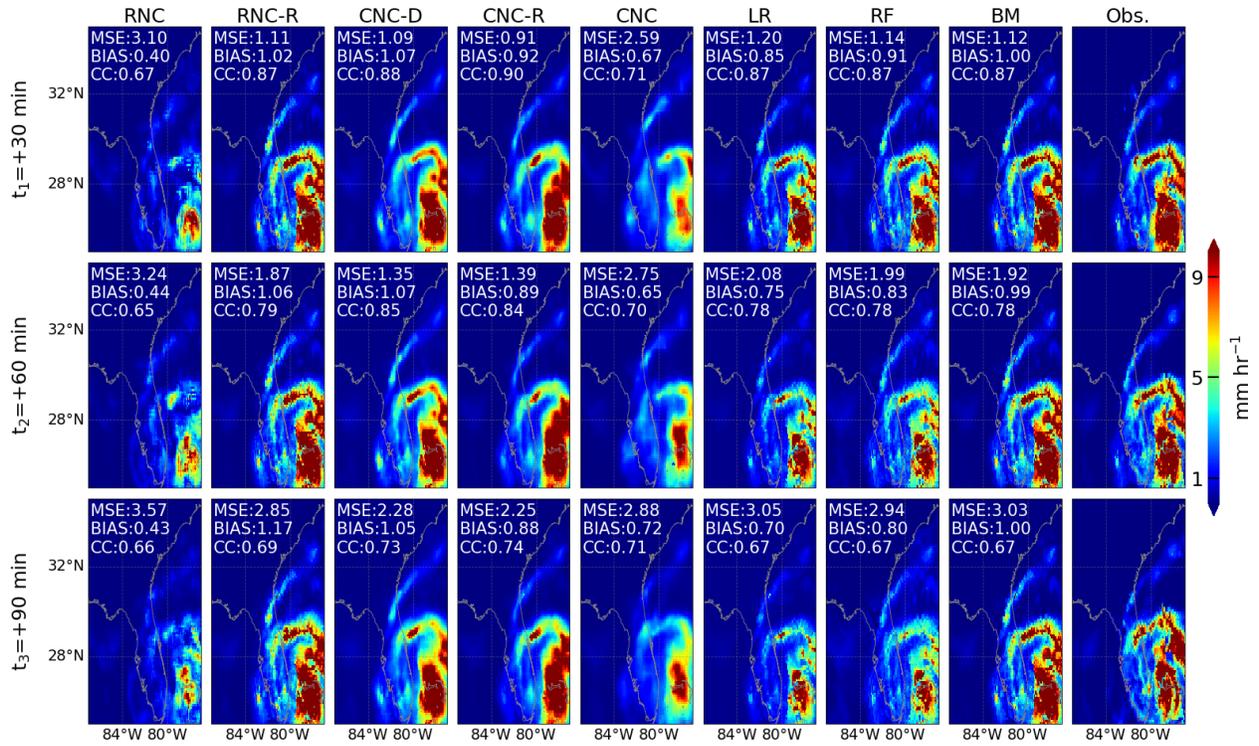

*Figure 9*: Precipitation nowcasting for category 1 Hurricane Fred (August 30, 2016) up to 90 minutes lead time using different models developed in this study. Each row shows a time step, and each column represents a model.

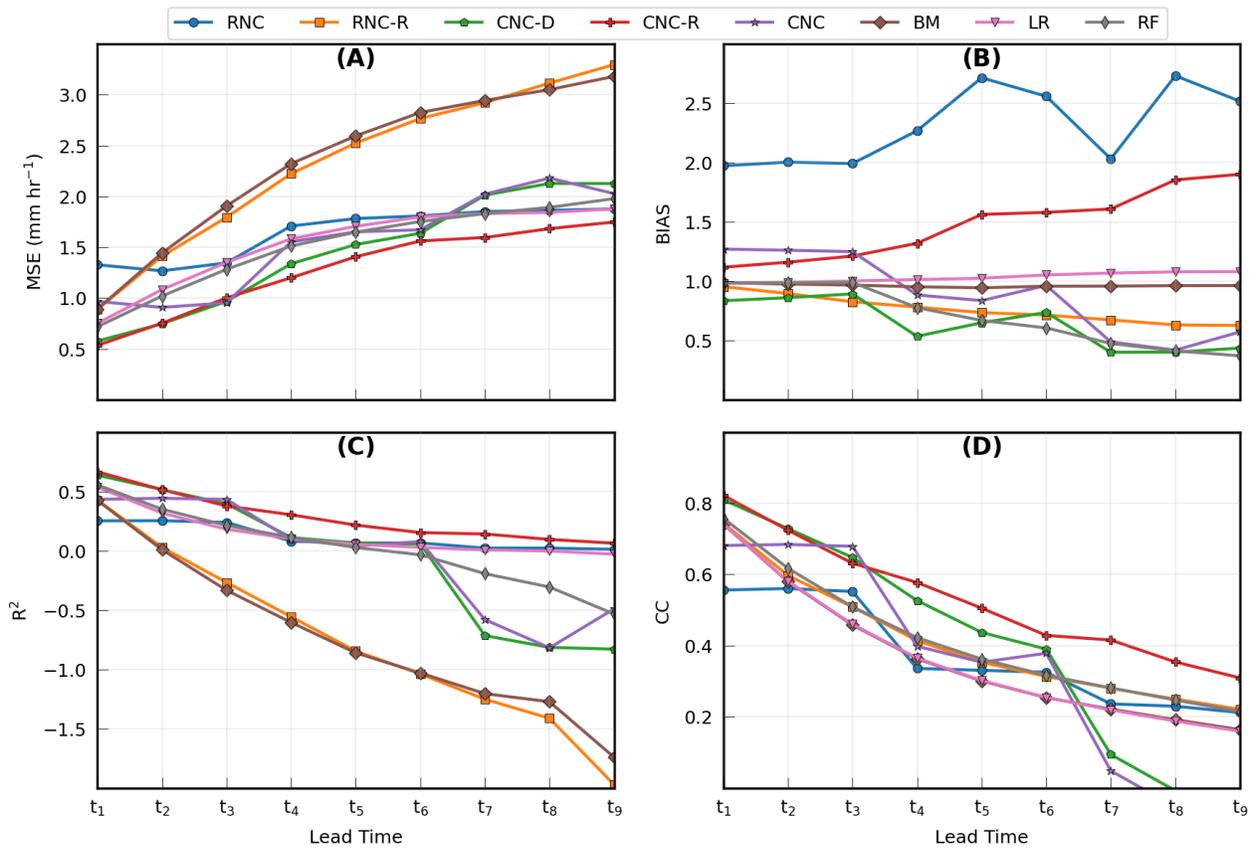

*Figure 10*: Mean performance of the models on the test set for the feedback loop experiment. Here, the lead time is increased up to 4.5 hours by recursively feeding the model its prediction three times.

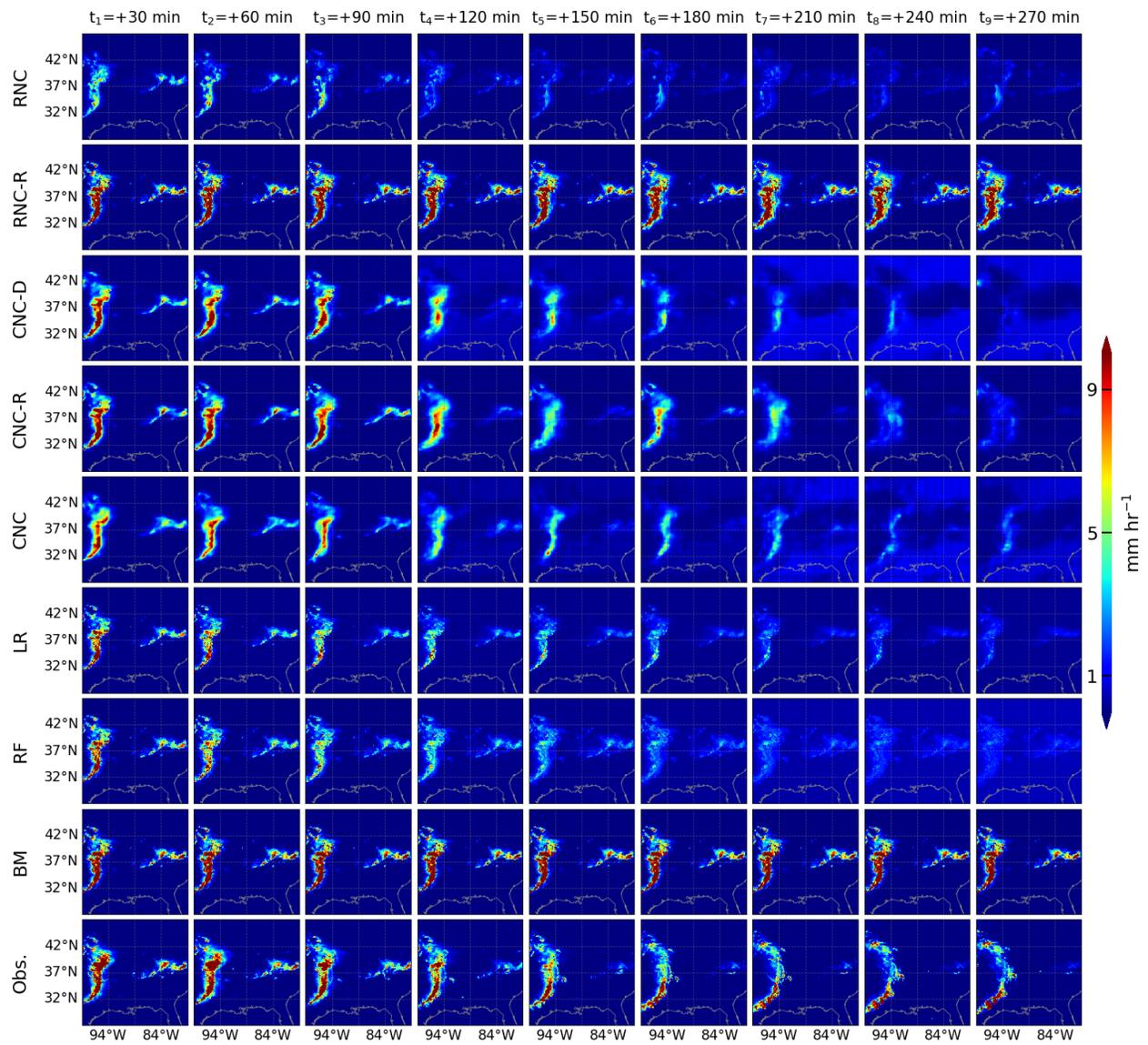

*Figure 11*: Performance of different models for the feedback loop experiment for an event on August 12, 2015. Each row shows a model, and each column represents a time step starting at 18:00 UTC.

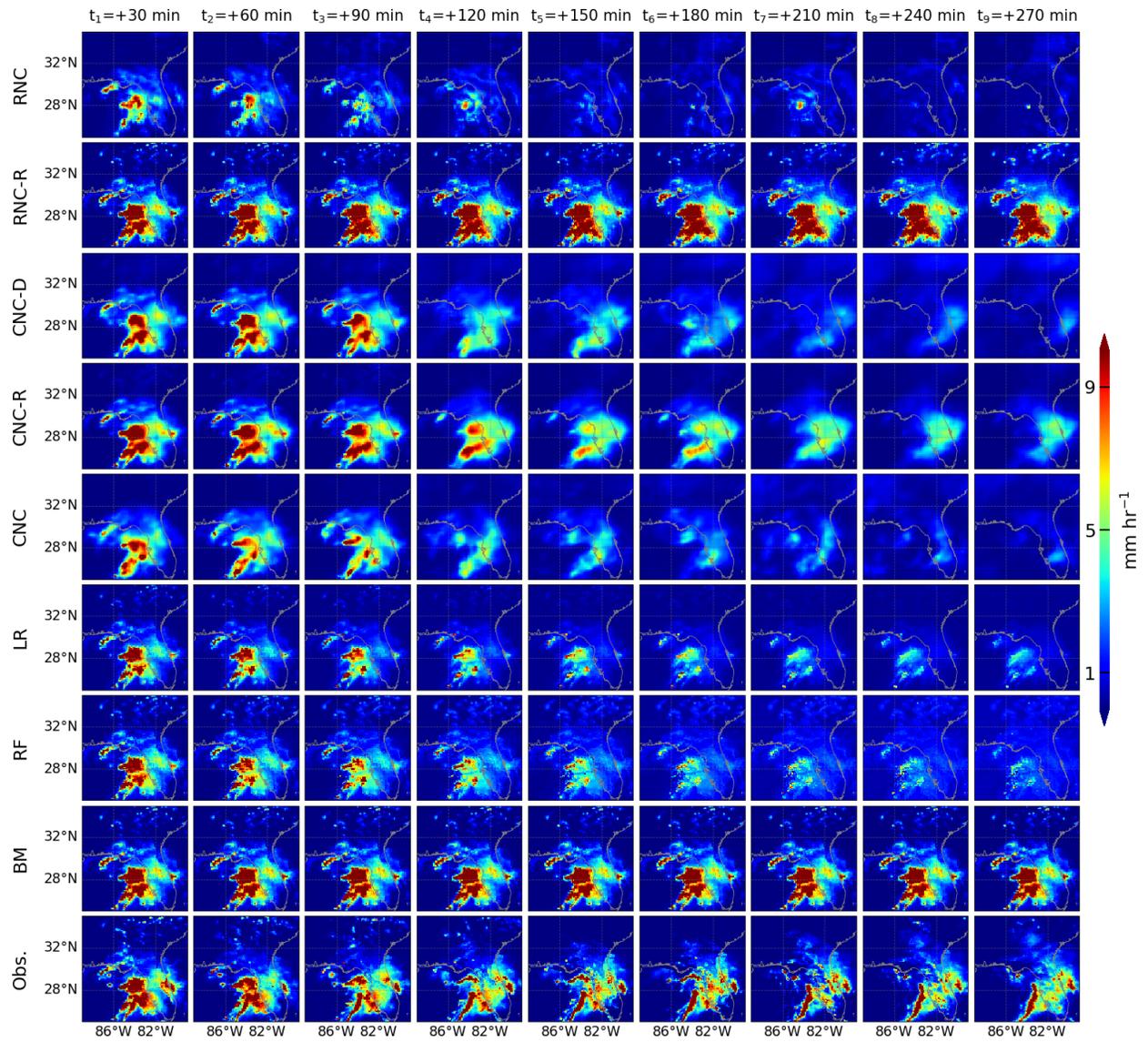

*Figure 12*: Performance of different models for the feedback loop experiment for an event on March 08, 2016. Each row shows a model, and each column represents a time step starting at 12:00 UTC.

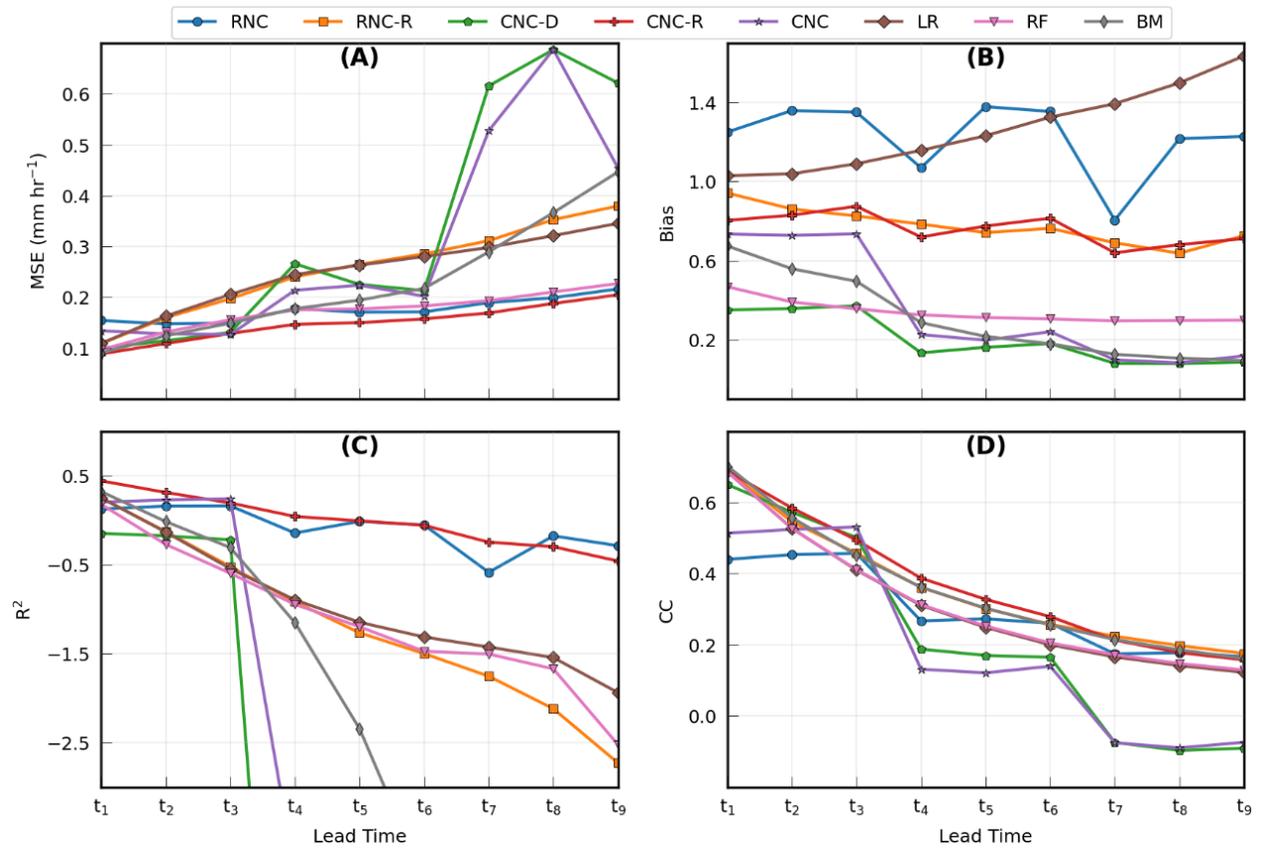

*Figure 13*: Performance of the models on the test set and using the feedback loop experiment over the Western CONUS (within the box shown in *Figure S1*).

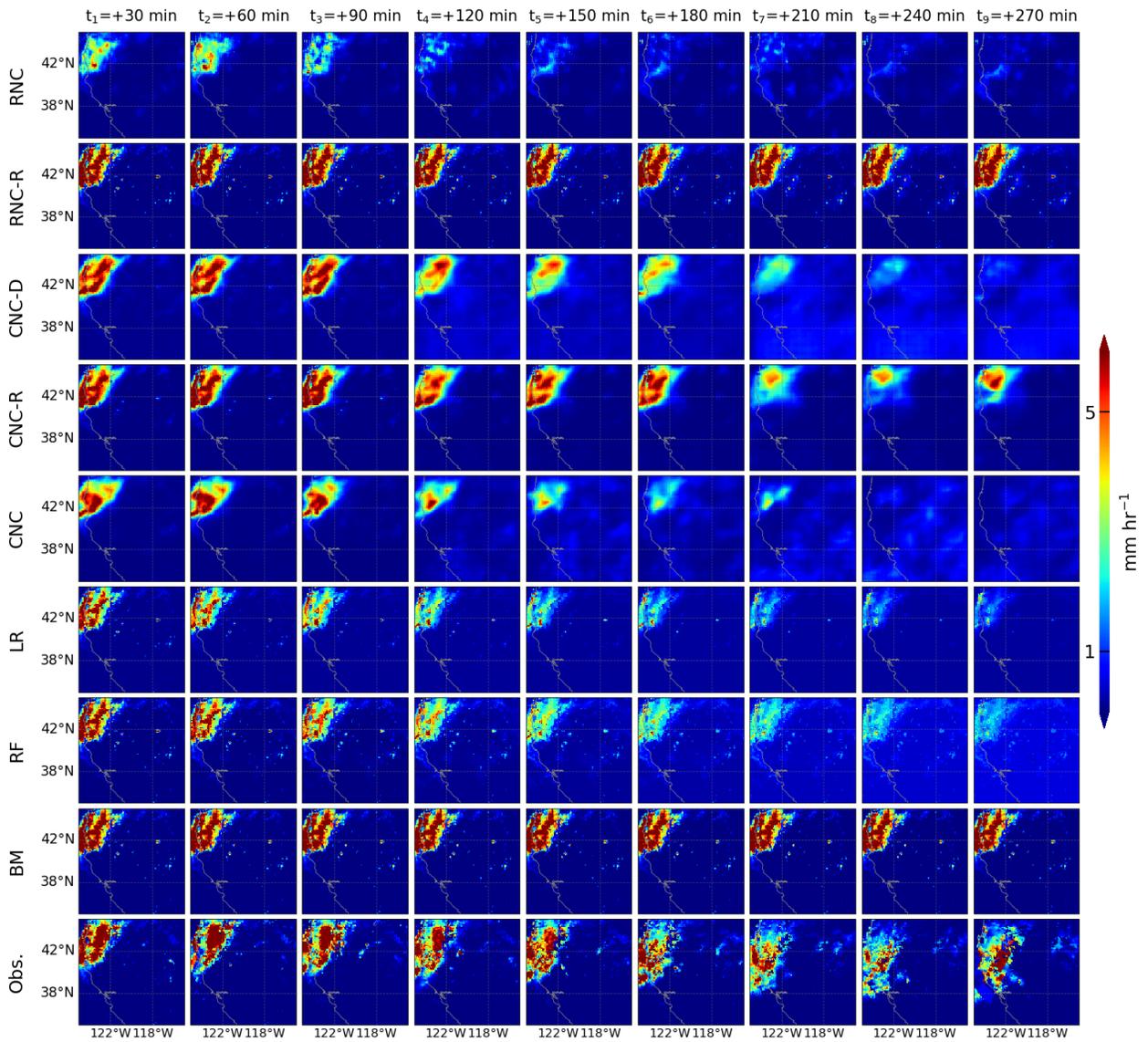

*Figure 14:* Performance of different models for the feedback loop experiment for a case study in Western CONUS on January 23, 2015, starting at 00:00 UTC. Each row shows a model, and each column represents a time step.

**Supplementary Figures**

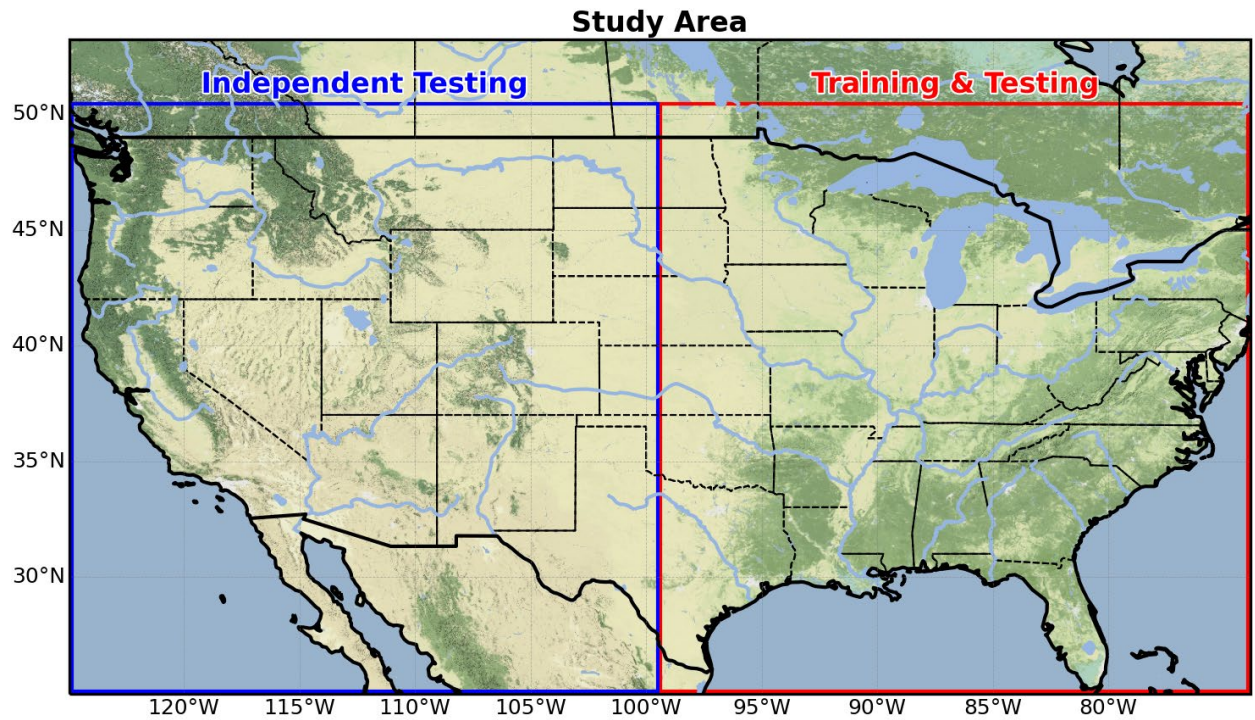

*Figure S1:* Study area: precipitation maps from the Eastern CONUS were used to train and test the models. The performance of the models was then further tested over the Western CONUS, data from which were not used to train and validate the models.

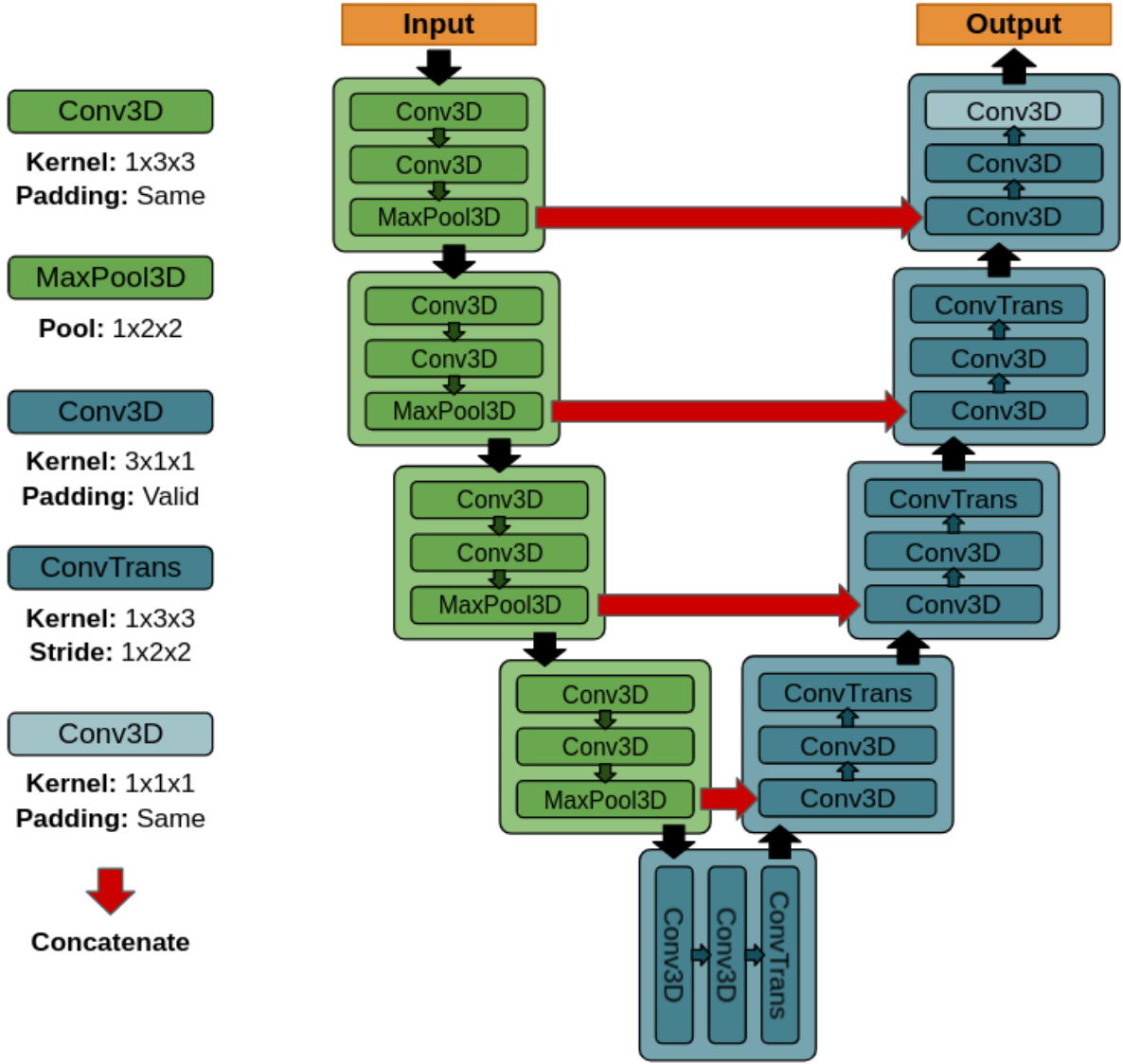

*Figure S2*: Architecture of the UNET model

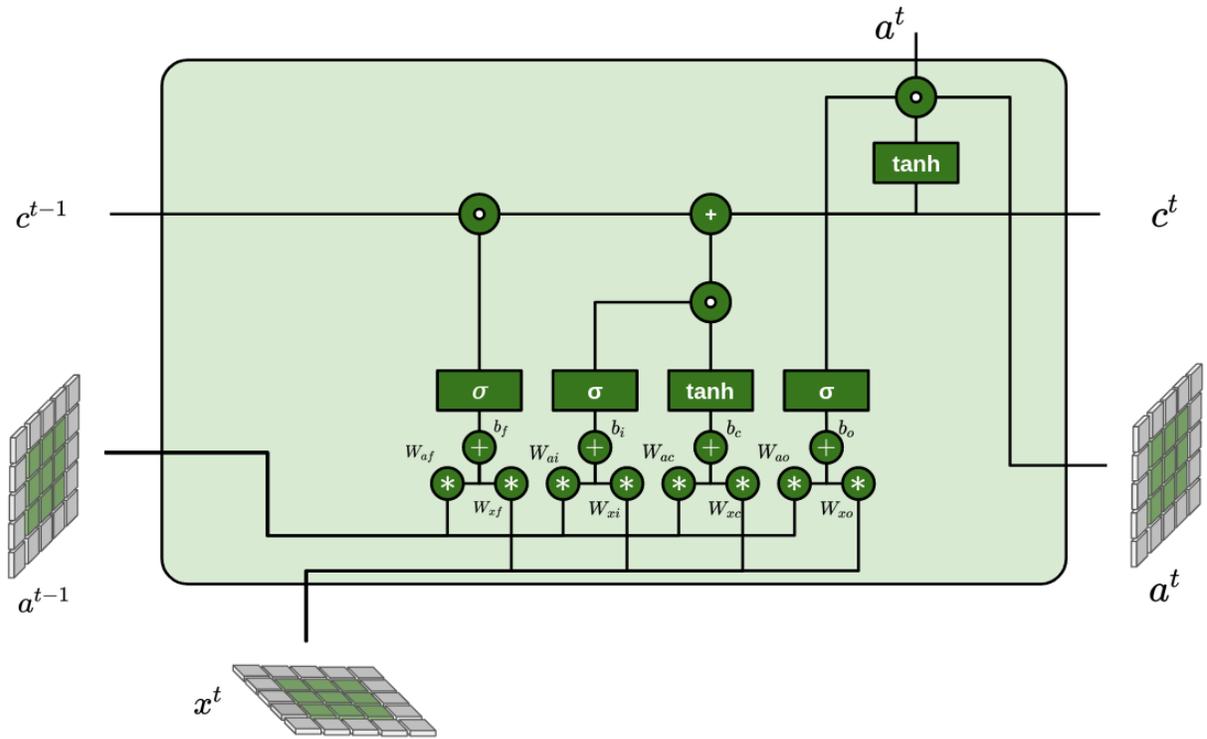

*Figure S3:* General architecture of *LSTM* and *ConvLSTM* networks.

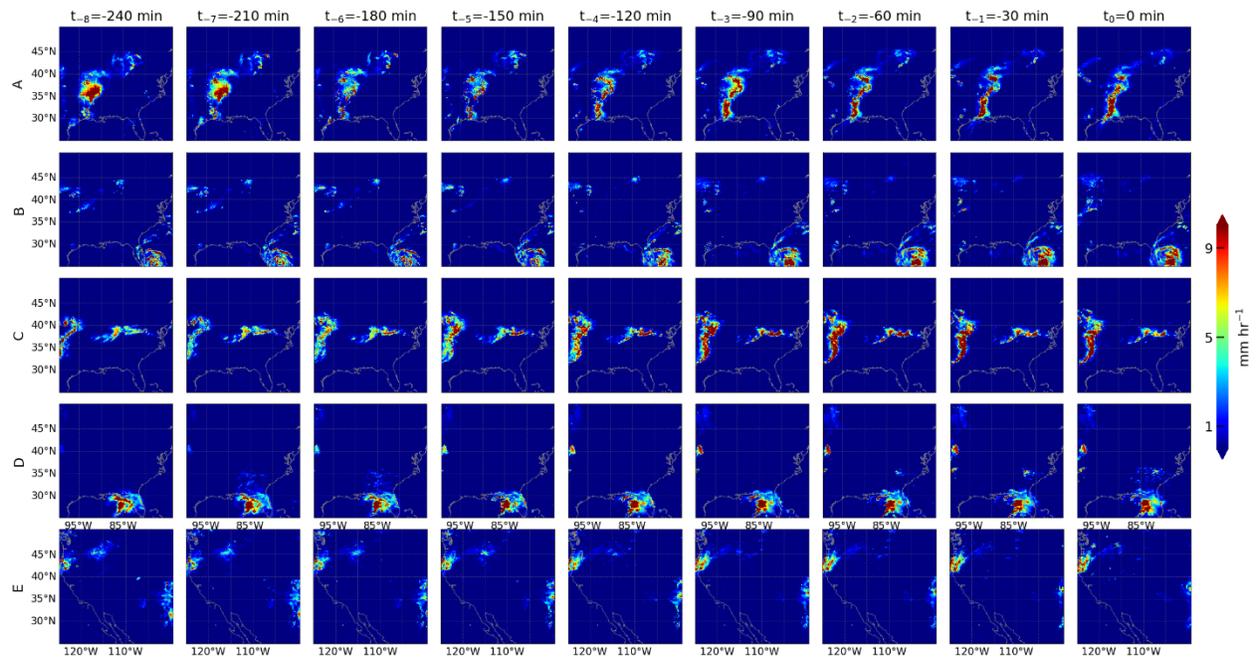

*Figure S4:* Input sequence for the different experiments in this study.